\documentclass[conference]{IEEEtran}
\usepackage{amsthm,amssymb,graphicx,multirow,amsmath,color,amsfonts}
\usepackage[update,prepend]{epstopdf}
\usepackage[noadjust]{cite}
\usepackage[latin1]{inputenc}
\usepackage{tikz}
\usepackage{bbm} 
\usepackage{pdfpages}
\usepackage{multirow}
\usepackage{comment}

\usepackage{algorithm}
\usepackage[noend]{algpseudocode}

\usepackage{balance}

\usepackage{optidef}
\usepackage{subfig}
\include{notation}
\allowdisplaybreaks 
\usepackage[colorinlistoftodos,bordercolor=orange,backgroundcolor=orange!20,linecolor=orange,textsize=scriptsize]{todonotes}

\usepackage{setspace}	

\graphicspath{{figures/}}

\newcommand{\algorithmicserver}{\textbf{Global Server}}
\algdef{SE}[SERVER]{Server}{EndServer}[1]{\algorithmicserver\ #1\ \algorithmicdo}{\algorithmicend\ \algorithmicserver}%
\algtext*{EndServer}
\newcommand{\algorithmicclient}{\textbf{Client i}}
\algdef{SE}[CLIENT]{Client}{EndClient}[1]{\algorithmicclient\ #1\ \algorithmicdo}{\algorithmicend\ \algorithmicclient}%
\algtext*{EndClient}
\makeatletter
\newcommand{\brokenline}[2][t]{\parbox[#1]{\dimexpr\linewidth-\ALG@thistlm}{\strut\raggedright #2\strut}}
\makeatother

\usepackage{romanbar}

\makeatletter
\renewcommand{\fnum@figure}{Figure \thefigure}
\makeatother

\makeatletter
\renewcommand{\fnum@table}{Table \thetable}
\makeatother

\usepackage{soul} 

\begin{document}
\title{
Joint Probability Selection and Power Allocation for Federated Learning
}

\author{
    \IEEEauthorblockN{Ouiame Marnissi, Hajar EL Hammouti, El Houcine Bergou
        }
    \IEEEauthorblockA{ School of Computer Science, Mohammed VI Polytechnic University, Ben Guerir, Morocco.\\
\{ouiame.marnissi, hajar.elhammouti, elhoucine.bergou\}@um6p.ma
}
}
\maketitle

\begin{abstract}













In this paper, we study the performance of federated learning over wireless networks, where devices with a limited energy budget train a machine learning model. The federated learning performance depends on the selection of the clients participating in the learning at each round. Most existing studies suggest deterministic approaches for the client selection, resulting in challenging optimization problems that are usually solved using heuristics, and therefore without guarantees on the quality of the final solution. We formulate a new probabilistic approach to jointly select clients and allocate power optimally so that the expected number of participating clients is maximized. To solve the problem, a new alternating algorithm is proposed, where at each step, the closed-form solutions for user selection probabilities and power allocations are obtained. Our numerical results show that the proposed approach achieves a significant performance in terms of energy consumption, completion time and accuracy as compared to the studied benchmarks.

\end{abstract}

\begin{IEEEkeywords}
Federated learning, Scheduling, Resource allocation, Non-iid data
\end{IEEEkeywords}
\section{Introduction} \label{sec:intro}
   In the last few years, the number of connected devices has increased significantly. Edge devices (e.g.,
smartphones, cameras, microphones, and house appliances) generate a large amount of data which is used to train machine learning (ML) models. However, to participate in ML training, devices need to share their personal information, often sensitive and private, with a centralized server. To overcome this issue, a novel ML paradigm has emerged, namely federated learning (FL)~ \cite{FL2017MC}. In FL, devices collaboratively train a shared ML model by iteratively sending ML parameters instead of their private data. As a consequence, the users' data is preserved and the computation load is shared between devices. However, implementing FL in real wireless networks comes with several key challenges. In fact, deep learning models train up to millions of parameters which leads to gigabytes of messages to be transmitted through limited-capacity networks. They also require intensive computation and communication operations which result in a high energy consumption. In this context, one may ask: can we devise a FL approach that encourages devices' participation in FL training while saving wireless resources and energy?


\subsection{Related Work} 

To reduce the communication costs, compression techniques were proposed. The objective of sparsification is to send a sparse vector of
the gradient estimates that includes only selected entries of the
vector. One way to sparsify the gradient vector is by keeping
the most impactful values of the gradient, and dropping the
others. Quantization consists of reducing
the number of digits that are used to encode the gradient vector. Another efficient FL technique suggests selecting a limited number of clients that participate in the FL task.  
Many works have been proposed to optimize resources while maintaining good performance of FL. Not only partial client
participation uses limited communication bandwidth, but when
optimally designed, it can also accelerate the FL convergence and minimize the computational resources. In general, a large number of participants per round speeds up the FL convergence~\cite{SGD2019Stich,fedavgnoniid2020,update2020amiri,3joint2021chen}. However, the limited wireless resources and energy restrict the number of users that can participate in each round. Therefore, it is important to schedule devices and optimize the wireless resources and energy accordingly. In~\cite{FEDCS2019nishio}, the authors present a protocol that maximizes the number of selected clients subject to a completion time constraint. They show through multiple simulation experiments that the proposed approach improves the FL convergence time. However, the proposed selection approach assumes that all devices are equally important, regardless of their local dataset size and distribution. Also, the proposed method may exclude some devices from participation only because their computation and communication costs are relatively high. This may lead to overfitting problems as some important data samples are not involved in the learning task.


To ensure some fairness among clients, authors in~\cite{robin2019yang} propose three scheduling policies, namely, random scheduling, round-robin, and proportional fairness. They compare their convergence rate and study their impact on the accuracy through theoretical analysis and simulation experiments. However, the study does not account for communication and computation energies. The trade-off between learning time and energy consumption is studied in~\cite{2tdoffs2019zomaya}. Other energy-efficient frameworks are proposed in \cite{energy2019zeng,energy2021yang} where the objective is to minimize the energy consumption for transmission in~\cite{energy2019zeng} and for both computation and transmission in \cite{energy2021yang}.

The authors of \cite{cvtime2020Chen} formulated an optimization problem that jointly considers user selection and resource allocation to minimize the FL convergence time. Since the studied problem is challenging, the authors decompose the optimization into two subproblems. The first one addresses the users' scheduling. In particular, a probabilistic approach based on the gradients' estimation is proposed. The second subproblem deals with resource allocation which is solved using an interior-point method. In the same context, the work in~\cite{joint2021shi} aims to minimize the loss function under energy and time constraints. The studied problem formulates the scheduling decision variables as binary integers and solves the problem approximately using constraint relaxation.  

In fact, resource allocation problems in the context of FL are often formulated as mixed integer non-linear programming, where the scheduling/selection parameters are considered as binary variables. As a consequence, the optimization problems are non-convex, most of the time NP-hard, and therefore, challenging to solve. Moreover, the proposed algorithms are heuristics with no performance guarantee and may require a large number of iterations to converge.

In our paper, we deal with this problem differently. We claim that, the binary association constraint can be relaxed. The association variables can be seen as probabilities. In particular, a user decides whether to participate in the FL training or not based on a probability distribution. As a consequence, instead of optimizing a deterministic objective function, the objective function is replaced by an expectation over the probability distribution. Therefore, since the devices' association is not deterministic (i.e., a device decides to send updates or not by randomly drawing samples from a probability distribution), it allows a large participation of devices, which leads to a certain fairness among devices.
It is to be noted that many works propose stochastic approaches for client selection, but the proposed approaches rely mainly on information on the loss and gradient values (e.g., selecting with probabilities proportional to the loss values or gradients' norms) and do not involve time and energy in their probabilistic selection methods.



\subsection{Contribution} In this paper we aim to speed up the convergence rate of FL within time and energy budgets. Our main contributions are summarized as follows. 
\begin{itemize}
    \item We formulate a joint scheduling and power allocation problem to maximize the weighted sum of the selection probabilities. We aim to optimize the probability of selection jointly with power allocation. In our problem formulation, we account for the transmission time along with the communication and computation energies. 
    
\item We propose an iterative algorithm to solve the optimization problem. At every step of the algorithm, closed-form solutions of user selection probabilities and power allocations are derived.
     
    \item Finally, we validate our selection approach with simulation experiments. In particular, we show that our approach outperforms uniform selection in terms of convergence time and consumed energy. We also show that under a highly-biased data setup, the probabilistic behavior of the proposed selection technique ensures the FL convergence. It allows the participation of a large number of devices as opposed to a deterministic version which restricts the selection to a fixed subset of devices, and therefore, prevents the FL from learning the global data pattern.
    
\end{itemize}
The remainder of this paper is organized as follows. The system model is described in section II. In section III, we formulate the studied problem as a joint power allocation and device selection problem, and we solve it in section IV. Simulation results are provided and analyzed in section V. Finally, section VI draws the conclusions of our paper.
\section{System Model}

\subsection{Communication Model}
We assume an FL setup where $N$ devices communicate with a parameter server over the wireless network. Suppose the clients use orthogonal frequency domain multiple access (OFDMA) for their transmissions. Each device $i$ transmits its gradients with a power $P_{ik}$ during time slot $k$. We also suppose that transmissions are attenuated with the distance. Hence, the received power of device $i$, at the server, is $P_{ik}d_i^{-2}$, where $d_i$ is the distance between device $i$ and the server. Each device $i$ is allocated a bandwidth $B_{i}$. Therefore, the achievable rate of device $i$ during time slot $k$ can be written 

\begin{equation*}
    r_{ik}(P_{ik})=B_{i}\log_2\left(1+\frac{P_{ik}d_i^{-2}}{\sigma^2}\right),
\end{equation*}
with $\sigma^2$ the power spectral density of the Gaussian noise. 
Therefore, When a device $i$ sends a gradient vector of size $S$ to the server during communication round $k$ (or equivalently time slot $k$), the transmission time $T_{ik}$ is required, where 
\begin{equation}
    T_{ik}(P_{ik})=\frac{S}{r_{ik}(P_{ik})}.
\end{equation}

\subsection{Machine Learning Model}
Each device trains the ML model on its local dataset. We denote by $D_i$ the dataset of device $i$, and we use $|D_i|$ to denote its size. To train the ML model, a loss function $F(\theta,\textbf{x},\textbf{y})$ is minimized 
\begin{equation}
    \min_{{\theta}}F({\theta},\textbf{x},\textbf{y})=\min_{{\theta}}  \sum\limits_{i=1}^{N} \alpha_i f_i({\theta},\textbf{x}_i,\textbf{y}_i),
\end{equation}
where $\alpha_i$ is the weight of local loss function $f_i$ of device $i$. ${\theta}$ is the ML parameter vector, ${\textbf{x}_i,\textbf{y}_i}$ corresponds to the pairs input-output for an ML model. In particular, $D_i=\{\textbf{x}_i,\textbf{y}_i\}$. The local loss function $f_i$ of device $i$ can be written

\begin{equation}
    f_i({\theta},\textbf{x}_i,\textbf{y}_i)=\frac{1}{|D_i|}\sum\limits_{l=1}^{|D_i|}f_{il}({\theta},\textbf{x}_{il},\textbf{y}_{il}),
\end{equation}
with $f_{il}({\theta},\textbf{x}_{il},\textbf{y}_{il})$ the loss of a sample $l$ at device $i$.

To learn the optimal parameters, a stochastic gradient descent (SGD) optimization is adopted. Therefore, at a communication round $k$, the server updates the model's parameters following the equation
\begin{equation}\label{updweights}
  \theta^{k+1}=\theta^{k}-\eta \sum \limits_{i \in \mathcal{S}^k}^{} \alpha_i \nabla f_i(\theta,\textbf{x}_i,\textbf{y}_i), 
\end{equation}
with  $\mathcal{S}^k$ the subset of clients that send their updates during communication round $k$, and $\nabla f_i(\theta,\textbf{x}_i,\textbf{y}_i)=\frac{\partial f_i}{\partial\theta} (\theta,\textbf{x}_i,\textbf{y}_i)$ is the gradient of the local loss function $f_i$.
\subsection{Energy Model}

To account for the energy consumption during the training, we consider both computation and communication energies. Let  $\gamma_i$ be the number of CPU cycles per second of client $i$. Let $C_i$ denote the number of CPU samples required to compute one sample of data. The computation energy of client $i$ is given by~\cite{compenergy2016mao}:

\begin{equation}
    E^c_i=\kappa C_i |D_i| \gamma_i^2,
\end{equation}
where $\kappa$ is the effective switched capacitance that depends on the hardware characteristics.   

To upload the model to the server, a device $i$ requires a communication energy $E_{ik}^{u}$ during communication round $k$
\begin{equation*}
 E_{ik}^u=P_{ik}T_{ik}(P_{ik}).   
\end{equation*}
Consequently, the total energy consumed by device $i$ during a communication round {k}, $E_{ik}$, is estimated as
\begin{equation}
    E_{ik}=E_{ik}^c+E_i^u.
\end{equation}

Due to the large bandwidth and transmit power of the server, the downlink communication time can be neglected. Hence, the energy of the model broadcast is ignored. Similarly, the computation energy of the server is neglected due to the limited computations performed at the server. Therefore, the consumed energy during one communication round is the sum of the computation and communication energies consumed by the selected devices. 

Let $a_{ik}$ be the probability that a client $i$ sends its updated parameters to the server during communication round $k$. Let $K$ be the maximum number of iterations (i.e., communication rounds) to achieve a target accuracy. To reduce the energy consumption, we assume that the expected consumed energy of device $i$ per each round $k$ does not exceed an energy budget of $E_i^{max}$. Furthermore, to speed up the convergence time, we assume that the expected transmission time of selected device $i$, $a_{ik}T_{ik}$ should not exceed a time threshold $\tau^{th}$. 

In the following, we formulate the joint probability selection and power allocation problem as a mathematical optimization, and propose an efficient algorithm to solve it. 
\section{Problem Formulation}

In a classical FL setup, a random subset of clients communicate their updates every communication round. However, the random selection does not provide the best performance in terms of convergence rate and energy efficiency. As a consequence, it is important to choose the selection probabilities in order to respect the energy budget of devices and improve the FL performance in terms of accuracy and convergence time.  Furthermore, since the clients transmit their gradients values over a resource-limited network, it is also crucial to optimally allocate the communication resources. Particularly, the transmit powers need to be optimized. 

Our goal is to maximize the expected weighted sum of the client selection probabilities. This objective follows the results in~\cite{FEDCS2019nishio} and \cite{SGD2019Stich} where it is shown that to speed up the FL convergence, it is important to select as many clients as possible during each communication round. Let $w_i$ be the weight of device $i$. The weights are included in the objective function to make it more general. In particular, in a heterogeneous setup where data is unequally distributed between devices, $w_i$ can be considered as $w_i=\frac{|D_i|}{\sum \limits_{j \in \mathcal{U}} |D_j|}$. We formulate the joint selection probability scheduling and power allocation for FL as follows:

\begin{maxi!}|s| 
{\textbf{a},\textbf{P}}{\sum\limits_{k \in \mathcal{K}}{{\sum \limits_{i \in \mathcal{U}} a_{ik}w_{i}\label{objective}}}}
{\label{GeneralOptimizationPb}}{}
\addConstraint{a_{ik}\left(P_{ik}T_{ik}(P_{ik})+E_i^c\right) \leq E_i^{\rm max} \quad\forall (\!i,k\!)\!\in\!\mathcal{U}\!\!\times\!\! \mathcal{K}\label{energy}}
{}{}
{}{}
\addConstraint{ \frac{a_{ik}S}{B_{i}\log_2\!\left(\!1\!+\!\frac{P_{ik}d_i^{-2}}{\sigma^2}\!\right)}\!\leq\!\tau^{th}\quad\forall (\!i,k\!)\!\in\!\mathcal{U}\!\!\times\!\! \mathcal{K}\label{temps}}
{}{}
\addConstraint{ 0\leq P_{ik} \leq P^{\rm max} \quad\forall (\!i,k\!)\!\in\!\mathcal{U}\!\!\times\!\! \mathcal{K}\label{pow}}
{}{}
\addConstraint{ 0\leq a_{ik} \leq 1 \quad\forall (\!i,k\!)\!\in\!\mathcal{U}\!\!\times\!\! \mathcal{K}\label{probuser}}
{}{}
\end{maxi!}


Constraint~(\ref{energy}) ensures that the energy budget of device $i$ for each round $k$ is respected. Constraint (\ref{temps}) ensures that the expected transmission time of device $i$ does not exceed a round time threshold. Finally, constraints (\ref{pow}) and (\ref{probuser}) define the optimization variables as continuous variables that belong to the intervals $[0,P^{\rm max}]$ for the power, and $[0,1]$ for the selection probabilities.

\section{Joint Probability Selection and Power Allocation Optimization}
Our optimization problem is hard to solve mainly due to the non convexity of constraints (\ref{energy}) and (\ref{temps}). Therefore, we propose an algorithm that iteratively solves problem (\ref{GeneralOptimizationPb}) through optimizing two subproblems, i.e., user allocation subproblem and power allocation subproblem. In particular, at each iteration, we first solve \textbf{P} for a fixed \textbf{a}, then the optimum \textbf{a} is updated based on the obtained value of \textbf{P}.
\subsection{Optimization with respect to power \textbf{P}}
Given a fixed user allocation \textbf{a}, problem (\ref{GeneralOptimizationPb}) becomes: 

\begin{maxi!}|s|
{\textbf{P}}{\sum\limits_{k \in \mathcal{K}}{{\sum \limits_{i \in \mathcal{U}} a_{ik}w_{i}\label{objectivepow}}}}
{\label{GeneralOptimizationpow}}{}
\addConstraint{\text{Constraints} (\ref{energy}),(\ref{temps}),(\ref{pow})}
{}{}
\end{maxi!}
Since the objective is a constant number, then the problem in (\ref{GeneralOptimizationpow}) is feasible if and only if the optimal $P_{ik}$ solution of this problem for a specific $i$ and $k$:
\begin{mini!}|s|
{P_{ik}}{\frac{a_{ik} P_{ik} S}{B_{i} log_2(1+\frac{P_{ik}}{d_i^2\sigma^2})} \label{objective_dink}}
{\label{GeneralOptimizationOrigin}}{}
\addConstraint{ P_{ik}^{\rm min}\leq P_{ik} \leq P^{\rm max} \label{pow_dink}}
{}{}
\end{mini!}
where
\begin{equation*}
    P_{ik}^{\rm min} = d_i^2\sigma^2(2^{\frac{a_{ik}S}{B_{i}\tau}}-1)
\end{equation*} satisfies the fact that the objective (\ref{objective_dink}) is lower than 
\begin{equation}\label{lowerbH}
    H_{ik} = E_i^{max} - a_{ik} E_i^c
\end{equation}
The objective has a fractional form and could be solved with limited complexity using Dinkelbach's algorithm \cite{DK1967}. More specifically, we reformulate the fractional form by decoupling the numerator and denominator whereby the joint optimization of both becomes easier. Therefore, let us consider the following problem
\begin{mini}|s|
{P_{ik}^{\rm min}\leq P_{ik}\leq P^{\rm max}}{a_{ik} P_{ik} S - \lambda B_{i} log_2(1+\frac{P_{ik}}{d_i^2\sigma^2})\label{objectivealone}}
{}{}
\end{mini}
For a fixed positive $\lambda$, problem in (\ref{objectivealone}) is convex and has the optimal solution of $P_{ik}^* = \frac{\lambda B_{i} }{a_{ik}Sln(2)}- d^2\sigma^2$ obtained by setting the first derivative of (\ref{objectivealone}) to $0$. We use Algorithm \ref{dinkel} for each $i$ and $k$ to obtain the optimal $\textbf{P}^*$ of problem (\ref{GeneralOptimizationpow}).

\begin{algorithm}
	\caption{Dinkelbach's Method} 
	\label{dinkel}
     \hspace*{\algorithmicindent} \textbf{Input:} $\lambda = \lambda ^0 > 0$ and $\epsilon > 0 $
	\begin{algorithmic}[1]
		 \For {$j=1,2,\ldots $ }
   \State Calculate the optimal $P_{ik}^* = \frac{\lambda^{j-1} B_{i} }{a_{ik}Sln(2)}- d^2\sigma^2$ of the problem \ref{objectivealone}
   \State \brokenline{Update $\lambda^{j} = \frac{a_{ik} P_{ik}^* S}{B_{i}log_2(1+\frac{P_{ik}^*}{d_i^2\sigma^2})} $}
   \If {$|\lambda^{j} - \lambda^{j-1}| < \epsilon$}
				\State \textbf{Output: $P_{ik}^*$}
				\EndIf
		\EndFor
	\end{algorithmic} 
\end{algorithm}
\subsection{Optimization with respect to user allocation \textbf{a}}
With given power $\textbf{P}$, the problem stated in (\ref{GeneralOptimizationPb}) becomes
\begin{maxi!}|s|
{\textbf{a}}{\sum\limits_{k \in \mathcal{K}}{{\sum \limits_{i \in \mathcal{U}} a_{ik}w_{i}\label{objective1}}}}
{\label{GeneralOptimizationuse}}{}
\addConstraint{\text{Constraints} (\ref{energy}),(\ref{temps}),(\ref{probuser})}
{}{}
\end{maxi!}
and, for each $i$ and $k$ has a solution of 
\begin{equation}\label{userallocform}
    a_{ik}^* = \min\left(1, \frac{\tau^{th}}{ST_{ik}(P_{ik})}, \frac{E_i^{\rm max}}{P_{ik}T_{ik}(P_{ik})+E_i^c}\right)
\end{equation}
\\
\begin{algorithm}
	\caption{Iterative algorithm} 
	\label{itera}
     \hspace*{\algorithmicindent} \textbf{Input:} {A feasible solution ({\bfseries a}$^0$, {\bfseries P}$^0$), $\epsilon > 0 $ and iteration number n = 0}
	\begin{algorithmic}[1]
  \Repeat 
    \State With given $\textbf{a}^n$ Solve the problem in (8) using Algorithm \ref{dinkel} and obtain the solution $\textbf{P}^{n+1}$
    \State With given $\textbf{P}^{n+1}$:
    \If{  $\textbf{P}^{n+1}$ makes (\ref{objective_dink}) lower bounded by $\textbf{H}$ defined in (\ref{lowerbH})}
    \State Obtain the solution $\textbf{a}^{n+1}$ using (\ref{userallocform})
    \Else 
    \State {Break}
    \EndIf
    \Until{$|(\sum\limits_{k \in \mathcal{K}}{{\sum \limits_{i \in \mathcal{U}} a_{ik}w_{i}}})^{n+1} - (\sum\limits_{k \in \mathcal{K}}{{\sum \limits_{i \in \mathcal{U}} a_{ik}w_{i}}})^{n}| < \epsilon $}
	\end{algorithmic} 
\end{algorithm}

Finally, Algorithm \ref{itera} gives the solution of the optimization problem stated in (\ref{GeneralOptimizationPb}) by iteratively solving problems in (\ref{GeneralOptimizationpow}) and (\ref{GeneralOptimizationuse}). At each iteration, the optimal solution of (\ref{GeneralOptimizationpow}) and (\ref{GeneralOptimizationuse}) is obtained and thus, the objective in (\ref{GeneralOptimizationPb}) is increased. Furthermore, since the user allocation is upper-bounded by $1$, the objective is also upper-bounded. Therefore, the convergence of the algorithm to a local optima is guaranteed.

In Algorithm \ref{algo}, we present the pseudo-code that describes our global FL approach.


\begin{algorithm}
	\caption{Joint Probability Selection and Power Allocation} 
	\label{algo}
     \hspace*{\algorithmicindent} \textbf{Input:} $N$ number of devices $i$, $\tau^{th}, E_i^{max}$\\
 \hspace*{\algorithmicindent} \textbf{Optimization:} 
 Solve the problem stated in (7) using Algorithm \ref{itera}
 to obtain the parameters $\textbf{a}^*$ and $\textbf{P}^*$.
 
  \hspace*{\algorithmicindent} \textbf{Learning:} 
	\begin{algorithmic}[1]
		 \For {$k=0,1,\ldots,K  $ communication rounds$ $ }
			\Server{}
				\State Send the global model $\theta^k$ to the clients 
				\State \brokenline{%
				Average the gradients of the devices }
                \State Update the global model as in equation (\ref{updweights})
            \EndServer
		    \Client{}	
			    \State Compute the gradient $g_i(\theta^k)$
				\State \brokenline{%
				Send gradient to the server with probability $a^*_{ik}$}
				\State \brokenline{%
				The vector is sent with power $P^*_{ik}$ over bandwidth $B_{i}$ }
			 \EndClient{}
		\EndFor
	\end{algorithmic} 
\end{algorithm}


In the next section, we show empirically the performance
of our joint client selection and power allocation strategy.
\section{Simulation Results}
In this section, we conduct experiments to test the performance of our approach. The performance is measured in terms of accuracy, completion time and consumed energy. In particular, we compare our selection strategy with three other selection approaches:
\begin{itemize}
 \item \textbf{Deterministic Selection}: This is the deterministic version of our proposed approach. In particular, the obtained probabilities are rounded up or down to illustrate a binary selection.
  \item \textbf{Uniform Selection \cite{FL2017MC}}: At each communication round, $M$ clients are selected uniformly at random to participate in the learning task. Note that, in general, this strategy does not necessarily satisfy the wireless and energy constraints of the problem.
  \item \textbf{Equally Weighted Selection~\cite{FEDCS2019nishio}}: This is the selection approach proposed in~\cite{FEDCS2019nishio}, where the selection variables are binary, and the objective function is the sum of equally weighed devices.
\end{itemize}
Due to the probabilistic behaviour of the uniform and probabilistic strategies, the results we present are averaged over $10$ different runs. 


\subsection{Simulation Setup}

Our experiments are conducted using Keras with Tensor-flow. We train a $3$ layers convolutional neural network (CNN) with $199,210$ parameters on a non-iid partitioned MNIST \cite{MNIST2010Lecun}; a dataset of hand-written numbers from $0$ to $9$.

We consider an area of $1 km^2$ where $100$ devices are scattered randomly and communicate with a server in the middle of the area. We assume a total bandwidth of $B = 10 MHz $ uniformly shared between the devices. The power spectral density of the Gaussian noise is equal to $\sigma^2$ = $10^{-12}$. We also assign for each device, a random energy budget $E_i^{bud}$ between $10^{-3}J$ and $100J$, that should not be exceeded in a round $k$. To assess the performance of our approach, we study two data biased scenarios. 
\begin{itemize}
  \item \textit{First scenario: Highly biased-data scenario}
 \end{itemize}
  Here, we consider a highly-biased data setup where each  device is missing some labels. In this scenario, we assume that the communication time threshold is $\tau^{th}=0.08 s$. We use Dirichlet distribution $Dir_K(\beta)$ to generate a skewed label distribution on devices \cite{diri@2021}. The concentration parameter $\beta > 0$ is used to control the degree of data imbalance level. A small $\beta$ implies large data heterogeneity. In this highly-biased data scenario, we choose a small value $\beta = 0.1$.  
 \begin{itemize} 
  \item \textit{Second scenario: Mildly-biased data scenario} 
\end{itemize} In this scenario, the Dirichlet parameter is higher compared to the first scenario, which results in a less biased data setup. In particular, $\beta = 0.3$. For this scenario, we consider that $\tau^{th}=0.5 s$. 
\subsection{Performance Evaluation}
\begin{itemize}
    \item \textit{Highly-biased data scenario}:
  \end{itemize}  
    
  \begin{figure}
\centering
{%
  \includegraphics[width=0.5\textwidth]{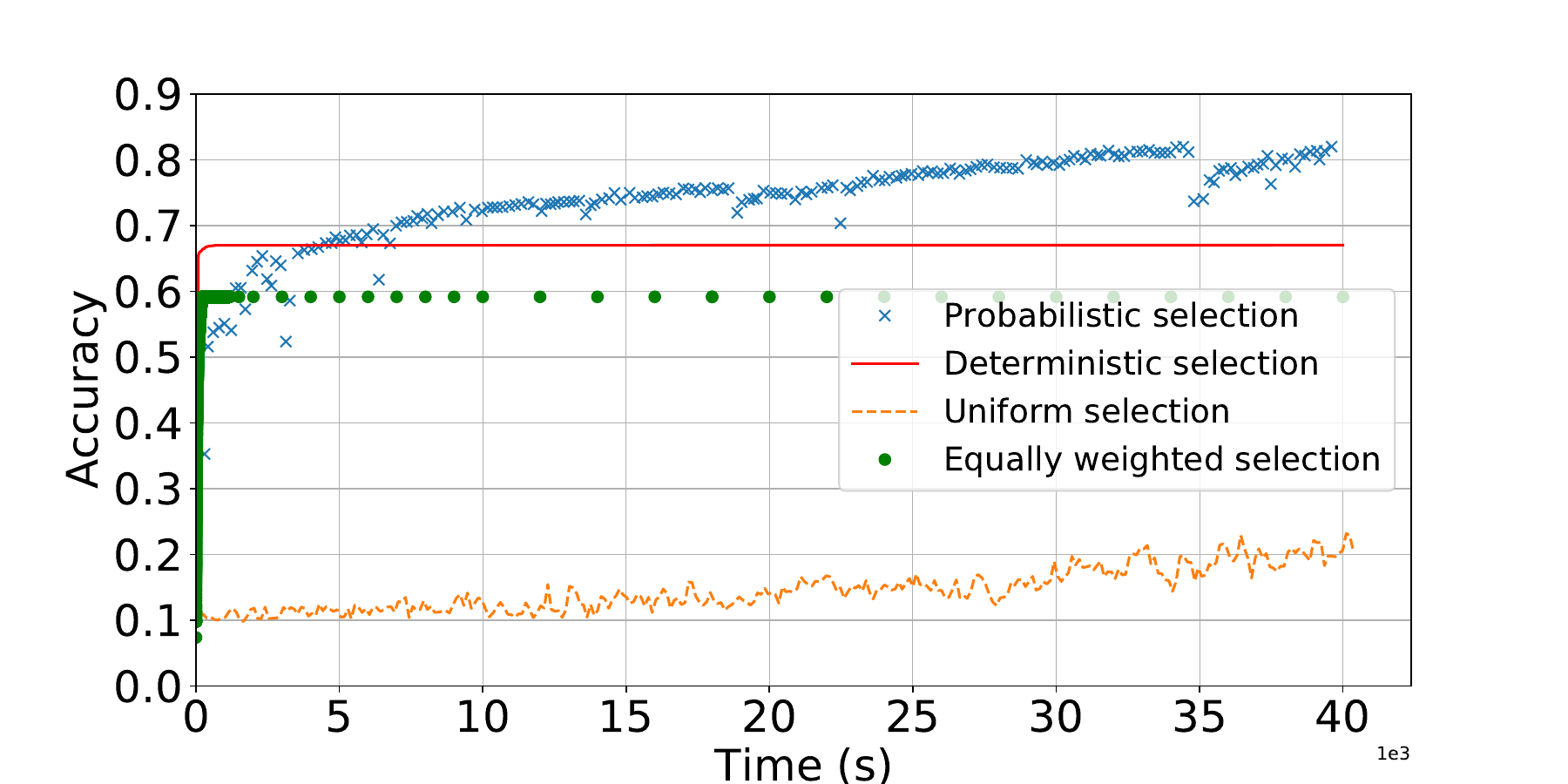}%
  }
\caption{Test accuracy for the highly-biased data scenario.}
\label{ACCUR1}
\end{figure}

 In Figure~\ref{ACCUR1}, we report the accuracy of the studied selection approaches against time. The communication time of each round corresponds to the transmission time of the stragglers. For this scenario, the proposed probabilistic selection outperforms the benchmarks in terms of accuracy. Although the expected number of selected devices per round is the same for probabilistic, deterministic and equally-weighted selections, the probabilistic property of our strategy allows the participation of different devices over iterations. In fact, even devices with a small selection probability participate in the learning task from time to time. This diversity in participation leads to high accuracy. On the contrary, the deterministic selection and equally-weighted selection achieve a lower accuracy. This can be clearly seen in Table \ref{time_acc} where the deterministic and equally-weighted selections never reach $80\%$. Finally, as it can be seen from Table \ref{time_acc}, the uniform method takes the longest time to reach a satisfying accuracy since it does not account for the energy and wireless constraints in the selection.

 Moreover, we calculate the total energy as the sum of the consumed energy per rounds. In Table~\ref{energy_acc}, we present the energy consumed to reach accuracies $59\%$ and $80\%$ respectively for each of the selection strategies. We can see that the uniform method consumes more energy than the other approaches.
\begin{table}[h]
\centering 
\begin{tabular}{l c c} 
\hline\hline 

\textbf{Achieved accuracy} & 59\% & 80\%\\
\hline 

Probabilistic selection & 1 307& 27 364\\
Deterministic selection & 31 & NA \\
Uniform selection & 80 113 & 126 747\\
Equally weighted selection &155 &NA\\[1ex]
\hline 
\end{tabular}
\caption{Time in (s) to achieve a target accuracy for the highly-biased data scenario.} 
\label{time_acc}
\end{table}

\begin{table}[h]
\centering 
\begin{tabular}{l cc} 
\hline\hline 
\textbf{Achieved accuracy} & 59\% & 80\%\\ 
\hline
Probabilistic selection & 625  & 13 061\\
Deterministic selection & 91 & NA \\
Uniform selection & 77 967 & 123 669\\
Equally weighted selection &463 & NA\\[1ex]
\hline 
\end{tabular}
\caption{Energy in (J) to achieve a target accuracy for the highly-biased data scenario} 
\label{energy_acc}
\end{table}
 \begin{itemize}
    \item \textit{Second scenario: Mildly-biased data scenario}
\end{itemize}
      \begin{figure}
\centering
{%
  \includegraphics[width=0.5\textwidth]{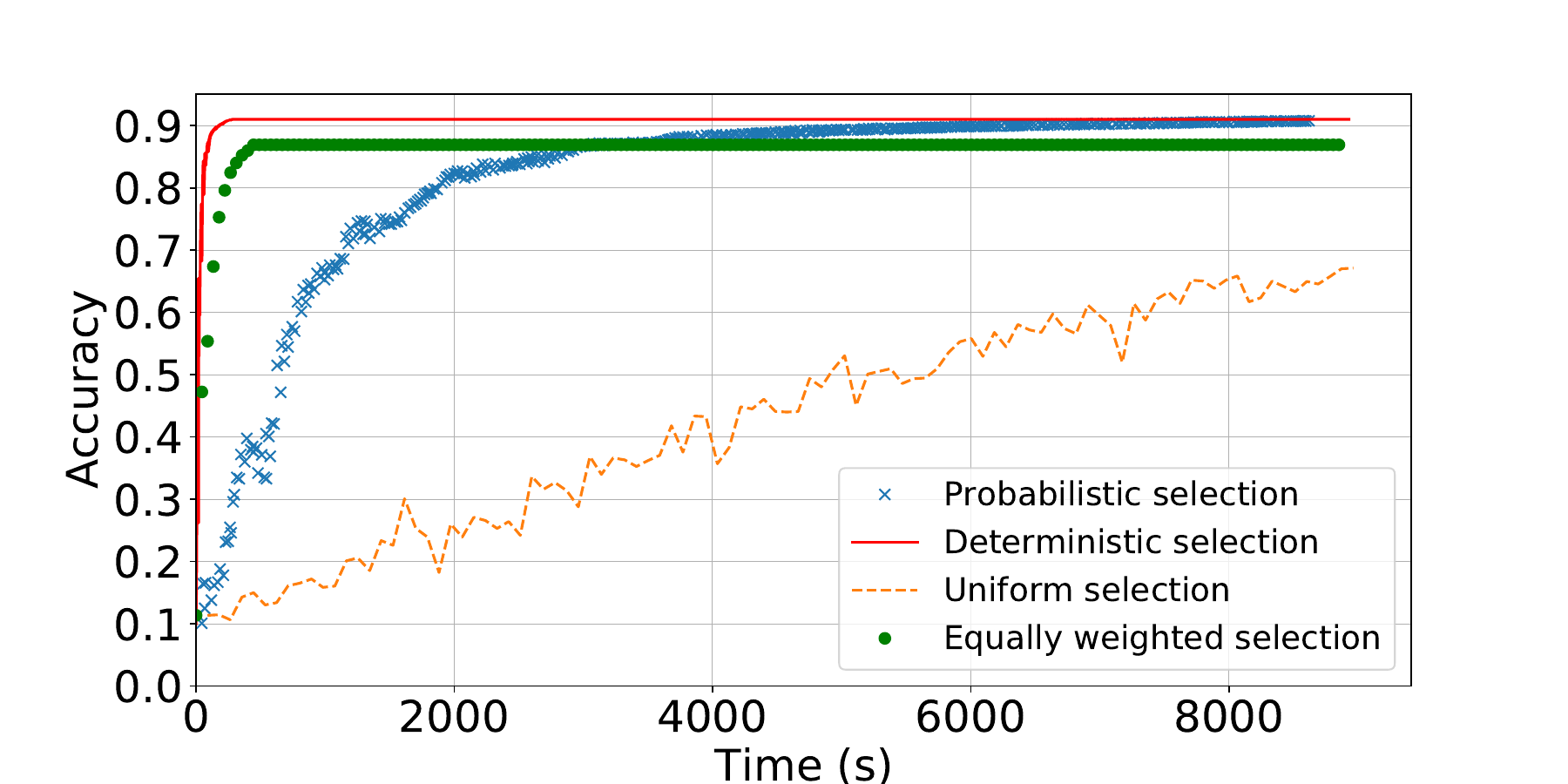}%
  }
\caption{Test accuracy for the mildly-biased data scenario.}
\label{ACCUR2}
\end{figure}

As it can be seen from Figure \ref{ACCUR2}, both probabilistic method and its deterministic version outperform the equally weighted selection. In fact, the proposed selection approach aims at maximizing the weighted sum of selection probabilities. Hence, devices with higher number of  weights are more likely to be selected. In our case, the weights are proportional to the local dataset sizes. As a consequence, devices with larger datasets have more chance to participate in the FL task. This improves the training time as larger number of samples is involved in the training. For the uniform approach, we observe the same behaviour as in the first scenario; it requires a long time to converge compared to the other approaches.

\begin{table}[h]
\centering 
\begin{tabular}{l c c} 
\hline\hline 
\textbf{Achieved accuracy} & 70\% & 86\%\\ 
\hline
Probabilistic selection & 1 145 & 2 834\\
Deterministic selection & 33 & 81\\
Uniform selection & 9 502 & 29 290\\
Equally weighted selection & 146 & 400\\[1ex]
\hline 
\end{tabular}
\caption{Time in (s) to achieve a target accuracy for the mildly-biased data scenario.} 
\label{time_acc2}
\end{table}

\begin{table}[h]
\centering 
\begin{tabular}{l cc} 
\hline\hline 
\textbf{Achieved accuracy} & 70\% & 86\%\\  \hline
Probabilistic selection & 591 & 1 438\\
Deterministic selection & 233 & 567 \\
Uniform selection & 29 225 & 90 348\\
Equally weighted selection &426 & 1 61\\[1ex]
\hline 
\end{tabular}
\caption{Energy in (J) to achieve a target accuracy for the mildly-biased data scenario.} 
\label{energy_acc2}
\end{table}
Table \ref{time_acc2} and Table \ref{energy_acc2} show the completion time and the energy consumption for the studied selection approaches to reach $70\%$ and $86\%$. The deterministic selection requires less time and consumes less energy as compared to its probabilistic version. This is mainly due to the reduced biased level of the studied scenario that does not require a large exploration of devices. The probabilistic version allows exploring more devices besides the most efficient ones, which may increase the energy consumption and the completion time. Finally, for the uniform selection, we observe the same results as in the first scenario.  \\
\section{Conclusion}
In this work, we have presented a joint probability selection and power allocation problem to maximize the weighted sum of selected devices while taking into account time and energy constraints. An iterative algorithm is presented to solve this problem, where at each step, closed-form solutions for user selection and power allocation are derived. Our numerical results showed that the
proposed approach achieves a significant performance in terms of
energy consumption, completion time and accuracy as compared
to the studied benchmarks. 
\balance
\bibliographystyle{IEEEbib}
\bibliography{references}


\end{document}